\documentclass[letterpaper, 10 pt, conference]{ieeeconf}  

\IEEEoverridecommandlockouts                              

\usepackage{graphics} 
\usepackage{epsfig} 
\usepackage{mathptmx} 
\usepackage{times} 
\usepackage{amsmath} 
\usepackage{amssymb}  
\usepackage{booktabs}
\usepackage{balance}
\usepackage{tikz}
\usepackage{atbegshi}

\title{\LARGE \bf
Scene-agnostic Hierarchical Bimanual Task Planning via Visual Affordance Reasoning 
}

\author{Kwang Bin Lee, Jiho Kang, and Sung-Hee Lee$^{*}$%
\thanks{The authors are with the Graduate School of Culture Technology, 
Korea Advanced Institute of Science and Technology (KAIST), 
291 Daehak-ro, Yuseong-gu, Daejeon 34141, Republic of Korea.
{\tt\small \{klee166, jhkang0408, sunghee.lee\}@kaist.ac.kr}}%
\thanks{$^{*}$Corresponding author}%
}

\begin{document}

\maketitle
\thispagestyle{empty}
\pagestyle{empty}

\begin{abstract}


Embodied agents operating in open environments must translate high-level instructions into grounded, executable behaviors, often requiring coordinated use of both hands. While recent foundation models offer strong semantic reasoning, existing robotic task planners remain predominantly unimanual and fail to address the spatial, geometric, and coordination challenges inherent to bimanual manipulation in scene-agnostic settings. We present a unified framework for scene-agnostic bimanual task planning that bridges high-level reasoning with 3D-grounded two-handed execution. Our approach integrates three key modules.
Visual Point Grounding (VPG) analyzes a single scene image to detect relevant objects and generate world-aligned interaction points. Bimanual Subgoal Planner (BSP) reasons over spatial adjacency and cross-object accessibility to produce compact, motion-neutralized subgoals that exploit opportunities for coordinated two-handed actions. Interaction-Point–Driven Bimanual Prompting (IPBP) binds these subgoals to a structured skill library, instantiating synchronized unimanual or bimanual action sequences that satisfy hand-state and affordance constraints. Together, these modules enable agents to plan semantically meaningful, physically feasible, and parallelizable two-handed behaviors in cluttered, previously unseen scenes. Experiments show that it produces coherent, feasible, and compact two-handed plans, and generalizes to cluttered scenes without retraining, demonstrating robust scene-agnostic affordance reasoning for bimanual tasks.
\end{abstract}

\newcommand\submittedtext{%
  \footnotesize This work has been submitted to the IEEE for possible publication. Copyright may be transferred without notice, after which this version may no longer be accessible.}

\newcommand\submittednotice{%
\begin{tikzpicture}[remember picture,overlay]
\node[anchor=south,yshift=10pt] at (current page.south) {\fbox{\parbox{\dimexpr0.65\textwidth-\fboxsep-\fboxrule\relax}{\submittedtext}}};
\end{tikzpicture}%
}


\maketitle
\submittednotice



\section{Introduction}

Embodied agents are increasingly expected to support everyday activities in open environments. These applications require agents to interpret high-level instructions, perceive 3D scene structure, and manipulate objects reliably across varied settings. At the core of this capability lies the task-planning problem: mapping a high-level instruction to a sequence of subgoals and actions that an agent can execute.

This problem becomes significantly harder in scene-agnostic contexts, where layouts, object configurations, and affordances vary widely. The planner must identify relevant objects, determine reachable interaction sites, and order actions while the agent navigates and manipulates objects in cluttered spaces. To act meaningfully under these conditions, agents must generate behaviors grounded in 3D scene geometry and aligned with the intended instruction.

Recent foundation models trained on large corpora, such as LLMs and VLMs, show strong multi-task generalization. These models can generate plausible action plans when conditioned on input prompts and have been adopted to inject semantic knowledge and commonsense reasoning into robotic task-planning pipelines~\cite{arajv2024Saynav, progprompt, Huang2022ZeroShotPlanners, zeng2022socraticmodels, wang2024llm3largelanguagemodelbasedtask}. However, most prior approaches plan tasks unimanually and rarely address the distinct challenges that arise when both hands must work together.

Scene-agnostic bimanual task planning maps a high-level natural language instruction to a sequence of coordinated left- and right-hand actions that complete a task more efficiently than a unimanual strategy. This involves reasoning about each hand’s current ownership and available actions; for example, a hand already grasping an object may place it into a trash bin while the other hand opens the lid, reducing the total number of steps. It also requires aligning actions with visual affordances in the 3D scene so that each hand interacts with the appropriate object or surface without interfering with the other. Coordination further depends on spatial arrangement: when two relevant objects lie within simultaneous reach, both hands may act together, whereas a larger separation may force them to operate independently. These tightly interdependent considerations distinguish bimanual planning from unimanual approaches and remain largely unresolved. To address this gap, we introduce situated awareness into a foundation LMM-based framework comprising three modules.

\textbf{Visual Point Grounding (VPG)} analyzes the 3D scene to ground interaction opportunities directly in its visual structure. It identifies relevant objects from an overview image, generates object-level object points for navigation and positioning, and extracts fine-grained interaction points. After lifting these points into world coordinates, VPG produces a unified representation of interaction sites that remains consistent across diverse, unseen scenes.

\textbf{Bimanual Subgoal Planner (BSP)} reasons over spatial adjacency and cross-object-point accessibility to produce subgoals that favor coordinated two-handed behavior. Using an object-point graph and a bimanual merge rule, it determines when object-oriented subgoals can be combined and when they must remain separate, reducing unnecessary serialization and aligning subgoals with spatial layout and two-handed reachability.

\textbf{Interaction-Point--Driven Bimanual Prompting (IPBP) } instantiates each subgoal as an executable two-handed motion sequence grounded in a structured skill library. Each skill specifies preconditions such as required hand states, object ownership, and affordance compatibility, along with synchronized unimanual or bimanual action patterns. By binding these constraints to VPG interaction points and current hand occupancy, IPBP produces action sequences that remain semantically consistent and physically feasible across scenes while enabling parallel, compatible hand actions whenever appropriate.

Together, these modules unify scene analysis, bimanual-aware subgoal structuring, and skill-grounded two-handed sequence generation into a coherent pipeline for bimanual planning in open-world 3D environments.

Our \textbf{contributions} are:
\begin{itemize}
    \item A unified framework for efficient bimanual planning in unstructured 3D scenes, handling evolving object states and multi-location two-handed interactions.
    \item A \textbf{Visual Point Grounding} module that detects task-relevant objects, generates object-level object points, and extracts affordance-aligned interaction points as world-grounded interaction sites.
    \item A \textbf{Bimanual Subgoal Planner} that improves bimanual efficiency through an object-point adjacency graph and a bimanual merge rule, producing compact, motion-neutralized subgoals aligned with spatial reachability.
    \item An \textbf{Interaction-Point--Driven Bimanual Prompting} module that retrieves skills with explicit preconditions and instantiates them using grounded interaction points and hand states to produce valid two-handed action sequences.
\end{itemize}

\section{Related Work}

\subsection{Task Planning}

Robotic planning has long represented tasks using symbolic languages such as PDDL and temporal logics~\cite{fox2003pddl2, emerson1990temporal}, paired with low-level motion planners to ensure geometric and kinematic feasibility~\cite{garrett2021integrated}. However, these pipelines depend on hand-engineered domains and carefully specified task descriptions, which limits their scalability across diverse environments. To reduce this burden, recent work leverages large language and vision–language models to translate under-specified natural language into structured planning goals, using commonsense reasoning to populate or adapt PDDL formulations~\cite{wang2024llm3largelanguagemodelbasedtask, xie2023translatingnaturallanguageplanning} that can be passed into TAMP frameworks~\cite{garrett2021integrated}, or into predefined symbolic skills executed by low-level policies~\cite{ahn2022can, Lin_2023}. Other approaches incorporate scene awareness by checking environment feedback, monitoring state changes, or using scene context to identify admissible actions at each step in simulators such as VirtualHome and ALFRED~\cite{Puig2018VirtualHome, ALFRED20, progprompt, Su2023SceneAware, Joublin_2024}. 

Although prior work captures high-level task semantics and decomposes them into symbolic programs, extending this process to the bimanual context remains an open problem.

\subsection{Visual Prompting in Motion Planning}
Visual prompting has emerged as an alternative way to ground task-level planning in visual observations. One line of work, such as VLM-TAMP~\cite{yang2024guidinglonghorizontaskmotion}, segments RGB scenes, tags objects with names, and prompts a VLM to generate symbolic subgoals that a TAMP~\cite{garrett2021integrated} pipeline can execute. Another line centers on the Set-of-Marks paradigm~\cite{yang2023setofmark}, which overlays colored marks or labels on images and prompts a VLM to identify or reason about the marked regions. This strategy has been applied to navigation framed as visual question answering~\cite{goetting2024endtoend} and to tabletop manipulation, where marked keypoints or affordances guide low-level policy learning or optimization objectives~\cite{google2024pivot, fangandliu2024moka, fang2025kalm}, and has recently been extended to the bimanual setting, as in ReKep~\cite{huang2024rekep}.

In this work, we address the broader problem of bimanual task planning across full-scene contexts that involve navigation, coordinated two-hand control, and hand-state–aware sequencing. 
\section{Method}

\subsection{Object Point and Interaction Point Representation}

Our framework models the scene using two complementary abstractions: \emph{object points} for navigation and positioning, and \emph{interaction points} for fine-grained manipulation. These components provide the spatial and semantic anchors used by downstream planning.

\subsubsection{Object Points}
An object point $o_i \in O$ denotes a task-relevant 3D location that identifies where the agent should navigate to interact with an object. Each object point is defined as
\[
o_i = (x_i, t_i), \qquad x_i \in \mathbb{R}^3,
\]
where $x_i$ is the 3D position and $t_i$ is the textual label indicating the object it belongs to. Object points form the search space for the Bimanual Subgoal Planner (BSP), which determines the order and grouping of object-level interaction regions.

\subsubsection{Interaction Points}
An interaction point $p_j \in P$ is defined as
\[
p_j = (x_j, d_j), \qquad x_j \in \mathbb{R}^3,
\]
where $x_j$ specifies a 3D contact location and $d_j$ is a semantic descriptor explained in Sec. \ref{sec:method_vpg}. 
This semantic package enables grounded bimanual skill reasoning and supports the Interaction-Point--Driven Bimanual Prompting (IPBP) module.

\subsection{Overview}

\begin{figure*}
  \centering
  \includegraphics[width=0.87\textwidth]{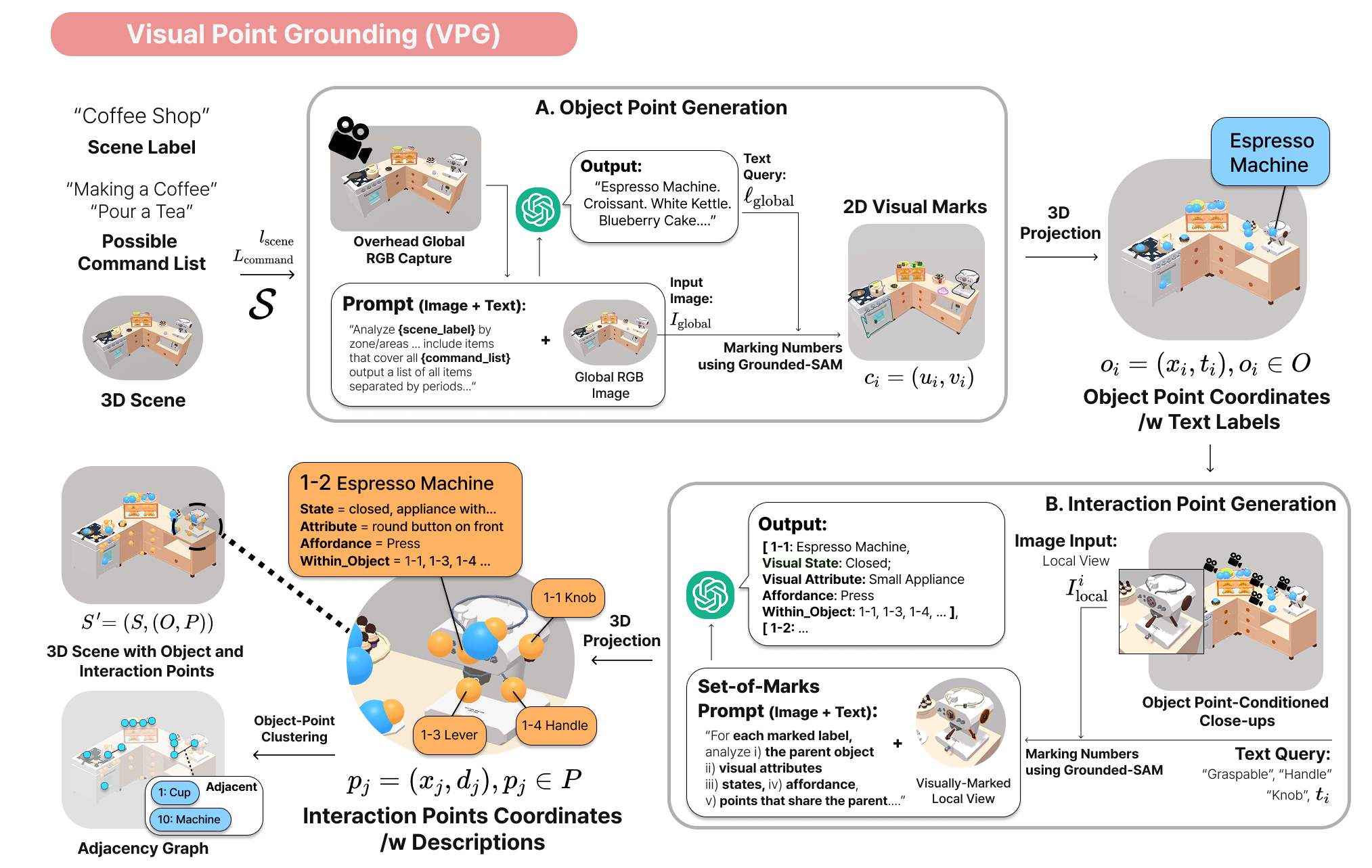}  
  \vspace{-5pt}
\caption{
Overview of the Visual Point Grounding (VPG) system.  
\textbf{A. Object Point Generation:} The system analyzes a global RGB overview 
image using an image–text prompt to identify instruction-relevant objects. 
Grounded-SAM marks each object in 2D, and these marks are lifted into 3D to 
create object points with associated labels.  
\textbf{B. Interaction Point Generation:} For each object point, a close-up 
local RGB view is processed with targeted queries (such as “handle,” “knob,” or 
“graspable”) to locate manipulable parts. Grounded-SAM and a Set-of-Marks 
prompt extract these part-level regions and project them into 3D as interaction 
points with descriptive attributes.  
\textbf{Scene Augmentation and Clustering:} The original scene is then 
augmented with all object and interaction points, and an adjacency graph is 
constructed by clustering nearby object points to capture local spatial 
relationships for downstream planning.
}
  \label{fig:overview_vpg}
  \vspace{-5pt}
\end{figure*}

Our framework consists of two stages. In the preprocessing stage, given a 3D scene $S$, 
a scene label $l_{\text{scene}}$, and a candidate command list $L_{\text{command}}$, 
the Visual Point Grounding Module constructs a grounded interpretation of the environment, 
as shown in Figure~\ref{fig:overview_vpg}.
\[
(O,P) 
= f_{\text{VPG}}(S,\, l_{\text{scene}},\, L_{\text{command}}).
\]
The scene label provides high-level contextual cues (e.g., ``Cafe,'' 
``Convenience Store''), while the command list helps ensure that potentially 
relevant objects are not overlooked.

\begin{figure*}[t]
  \centering
\includegraphics[width=0.87\textwidth]{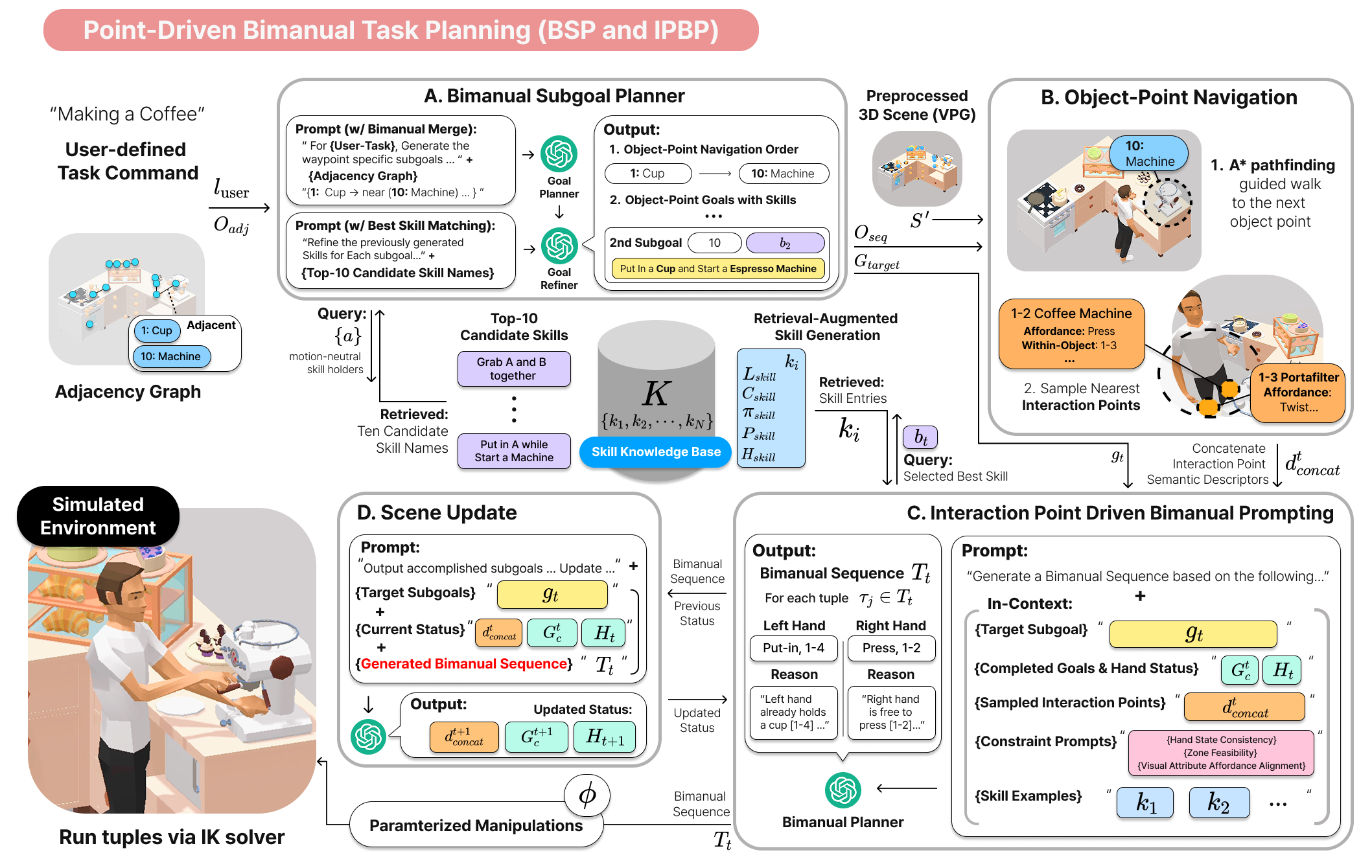}
  \vspace{-5pt}
\caption{
Overview of the Point-Driven Bimanual Planning system, composed of the 
Bimanual Subgoal Planner (BSP) and the Interaction-Point–Driven Bimanual 
Prompting module (IPBP).  
\textbf{A. Bimanual Subgoal Planner:} Using the user's task command together 
with the adjacency graph produced by VPG, BSP selects task-relevant object 
regions and generates a sequence of bimanual subgoals, refined through 
skill-name matching and retrieval from the skill knowledge base.  
\textbf{B. Object-Point Navigation:} The agent walks through the pre-processed 
scene produced by VPG, navigates to each subgoal’s object point using A* 
pathfinding, and samples the nearest interaction points visible from that 
location.  
\textbf{C. Interaction-Point–Driven Bimanual Prompting:} IPBP combines the 
refined skill, the current subgoal, sampled interaction points, and the 
agent’s hand state to produce synchronized bimanual action tuples guided by 
retrieved coordination patterns.  
\textbf{D. Scene Update:} Executed action tuples update the hand state, object 
interactions, and remaining subgoals, enabling iterative execution of the full 
bimanual manipulation sequence.
}
\vspace{-5.3pt}
  \label{fig:overview_planning}
\end{figure*}

Given the grounded scene interpretation $(O, P)$ and a user-specified task 
$l_{\text{user}}$, the system generates a complete bimanual plan aligned with 
object points and interaction points, as shown in 
Figure~\ref{fig:overview_planning}. The Bimanual Subgoal Planner (BSP) first 
constructs a structured task skeleton consistent with scene geometry and two-handed 
reachability:
\[
(O_{\text{seq}},\, G_{\text{target}},\, L_{\text{best\_skill}})
= f_{\text{BSP}}(l_{\text{user}},\, O),
\]
where $O_{\text{seq}}$ is the ordered list of object points to visit, 
$G_{\text{target}}$ specifies the subgoals at those points, and $L_{\text{best\_skill}}$ 
selects an appropriate skill for each subgoal. This stage determines where to act 
and what must be achieved at each location.

For each object point $o_t$ in $O_{\text{seq}}$ with corresponding subgoal $g_t \in G_{\text{target}}$ 
and skill $b_t \in L_{\text{best\_skill}}$, the Interaction-Point--Driven Bimanual Prompting module (IPBP) 
produces a synchronized sequence of bimanual action tuples:
\[
T_t = f_{\text{IPBP}}(o_t,\, g_t,\, b_t,\, H_t,\, d_{\text{concat}}^{(t)}),
\]
where $H_t$ is the current hand state and $d_{\text{concat}}^{(t)}$ is a text
summary capturing all feasible contact options at $o_t$.

\begin{figure*}[t]
    \centering    \includegraphics[width=0.77\textwidth]{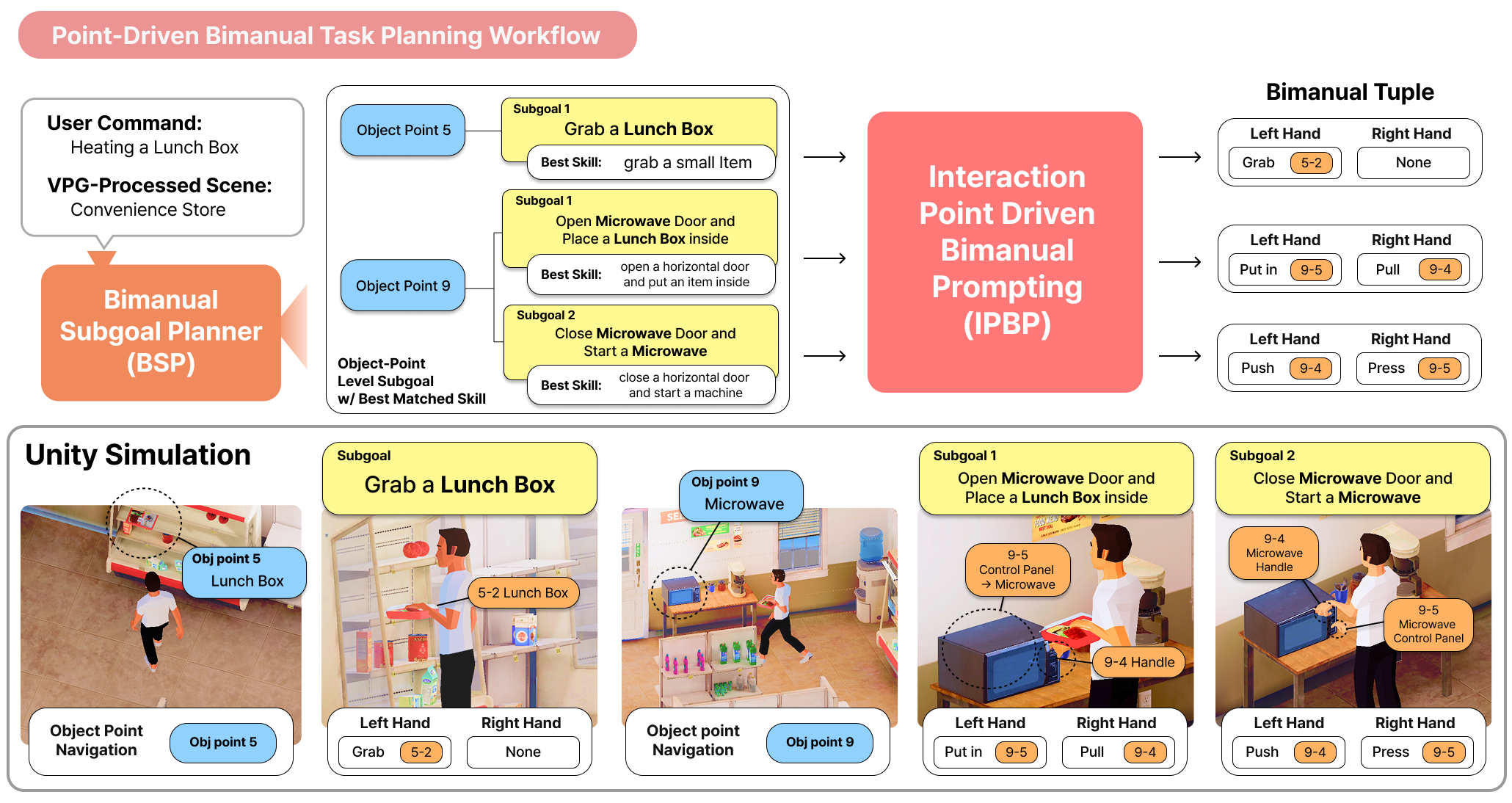}
    \caption{Workflow of point-driven bimanual task planning. Given a high-level user command (heating a lunch box) and a VPG-processed scene (convenience store), the Bimanual Subgoal Planner (BSP) forms object-point–level subgoals and assigns the most suitable canonical manipulation skills. Each subgoal is then converted into grounded bimanual tuples by the Interaction Point Driven Bimanual Prompting module (IPBP), which binds manipulation templates to the retrieved interaction points and current hand states. The resulting action sequence is executed in Unity, where the agent navigates to each object point and performs the required two-handed interactions, producing coherent, feasible, and visually grounded bimanual behavior.}
    \label{fig:workflow}
    \vspace{-5.5pt}
\end{figure*}

The final output of the task planner is the ordered sequence
\[
\Pi = \{T_1,\, T_2,\, \dots,\, T_k\},
\]
forming a spatially grounded, coordination-aware bimanual task plan for the user-defined task. 
Figure \ref{fig:workflow} illustrates a concrete example of task planning through BSP and IPBP.
Details of each module are provided next.

\subsection{Visual Point Grounding (VPG)}
\label{sec:method_vpg}

\textbf{Object Point Generation.}
Given a 3D scene $S$, a scene label $l_{\text{scene}}$, a command list 
$L_{\text{command}}$, and a single high-angle RGB overview image 
$I_{\text{global}}$, a VLM is prompted to identify object categories in the 
global view, producing a set of object labels $\ell_{\text{global}}$ guided by 
the scene and command cues. Grounded-SAM \cite{ren2024grounded} then localizes 
each identified object by segmenting its corresponding region in the image and 
attaching the label $t_i \in \ell_{\text{global}}$. The center $(u_i, v_i)$ of 
each segmented region is projected into 3D via raycasting, yielding an object 
point $o_i$. The collection of all such points forms the object-point set $O$, 
which provides the navigation anchors used by downstream planning modules.

Subsequently, we construct an adjacency graph $O_{\text{adj}}$ by linking pairs of
object points whose 3D positions fall within a predefined distance threshold.
These adjacency relationships capture which object points are jointly reachable
by the agent’s two hands, forming the structural basis for identifying feasible
bimanual opportunities.

\textbf{Interaction Point Generation.}
For each object point $o_i$, the system captures a close-up local RGB image 
$I_i^{\text{local}}$. Grounded-SAM is applied using a fixed set of 
affordance-oriented prompts—``graspable,'' ``handle,'' and ``knob''—together 
with the object’s textual label $t_i$ to ensure part detection even when visual 
cues are subtle. All detected regions are then annotated using a Set-of-Marks \cite{yang2023setofmark} 
prompt, which prompts the VLM to generate the interaction 
semantic descriptor $d_j$. 
The descriptor includes:  
\begin{itemize}
    \item Visual attributes relevant to manipulation (e.g., \textit{wall-mounted}, \textit{hinged})
    \item Object part labels (e.g., handle, spout, lid, button)
    \item Appearance-driven affordance cues (e.g., pressable, graspable, rotatable)
    \item The numeric IDs of other interaction points that belong to the same parent object.
\end{itemize}

Each 2D centroid is lifted into 3D to form an 
interaction point $p_j = (x_j, d_j)$, and the collection of these points forms 
the interaction-point set $P$ used for fine-grained affordance reasoning.

\subsection{Bimanual Subgoal Planning (BSP)}

BSP operates in two stages: a goal planner first generates object-point–specific subgoals and a goal 
refiner then specializes these subgoals by selecting the most appropriate 
skill name from the skill library.

\subsubsection{Initial Subgoal Generation and Bimanual Merging}
Using $O_{\text{adj}}$ and the user-given task command, the 
goal planner produces  
(1) an ordered sequence of object-point IDs (e.g., $3 \rightarrow 4$) specifying  
the spatial visitation order;  
(2) a sequence of free-form subgoal descriptions for each object point 
(e.g., at point~3: ``grab the lunch box''; at point~4: 
``open the microwave door and place the lunch box inside''); and  
(3) a corresponding sequence of motion-neutral \emph{abstract free-form skills}, 
representing possible bimanual operations required to satisfy each subgoal.

Together, these outputs form an index-aligned triplet representation
\[
(o_i,\; g_i,\; a_i), \qquad i = 1,\dots,n.
\]

The planner then applies a bimanual merge rule that determines when two nearby 
objects can be jointly reached and manipulated from a single stance.  
Two subgoals are merged when their object points are simultaneously reachable 
and their abstract skills $a_i$ can be executed in parallel without violating 
the hand-state or affordance constraints encoded in $O_{\text{adj}}$.  
For example, if a cup and a machine button are both within reach, the subgoals 
``grasp the cup'' and ``press the button'' may be merged into a single bimanual 
subgoal anchored at the point that supports joint reachability.

The merge rule also resolves continuity expressions such as ``while holding~[A].''  
If an earlier subgoal already establishes a grasp of object~[A], then any later 
subgoal or abstract skill that references ``while holding~[A]'' is merged back 
into that earlier step, ensuring the final subgoal sequence maintains explicit, 
continuous hand--object relationships.

\subsubsection{Refinement via Best Skill Matching}
The triplet $(o_i, g_i, a_i)$ is then passed to the goal refiner.  
Because the free-form skill $a_i$ generated by the planner may be ambiguous, overly abstract, or not directly supported by the predefined skill set, the refiner retrieves 10 candidate canonical skill names from $L_{\text{skill}}$ in the skill knowledge base $K$ (described later in Section~E) by embedding $a_i$ with the Sentence Transformer model \texttt{all-mpnet-base-v2} \cite{reimers-2020-sentence-transformers} and selecting the most semantically compatible candidates in the skill embedding space.
From the retrieved candidates, the refiner selects a single best skill $b_i$ 
that is  
(1) feasible at the object point $o_i$,  
(2) semantically compatible with $a_i$ 
, and  
(3) executable under the scene context and the adjacency constraints encoded in 
$O_{\text{adj}}$.

This produces an index-aligned sequence of refined skill assignments.  
Thus, the output of BSP is the aligned triplet
\[
(O_{\text{seq}},\; G_{\text{target}},\; L_{\text{best\_skill}}),
\]
where $O_{\text{seq}} = (o_1,\dots,o_n)$ is the ordered object-point sequence,  
$G_{\text{target}} = (g_1,\dots,g_n)$ is the sequence of subgoals, and  
$L_{\text{best\_skill}} = (b_1,\dots,b_n)$ contains the finalized best-skill 
assignments.

\subsection{Interaction-Point--Driven Bimanual Prompting (IPBP)}

Before sampling interaction points, the agent navigates to each target object point $o_t$ using A* pathfinding in the preprocessed 3D scene. Once it reaches $o_t$ and aligns its stance, the system identifies all interaction points that are reachable from that location.

From this set, IPBP extracts the interaction points within a predefined 
distance threshold to form $P^{(t)} \subseteq P$, and assigns each interaction point a \textit{left}, \textit{mid}, or 
\textit{right} spatial zone based on its 3D position relative to the agent. Each interaction point $p$ is represented by its 
descriptor $d(p)$ augmented with this zone label. These augmented descriptors 
are concatenated into
\[
d_{\text{concat}}^{(t)} 
= \text{concat}\,\{\, d(p)\ \text{with zone label} \mid p \in P^{(t)} \}.
\]

\subsubsection{Bimanual Tuple Generation}

IPBP generates a synchronized bimanual action sequence that satisfies geometric, affordance, and 
hand-state constraints given $(o_t,\, g_t,\, b_t,\, H_t,\, d_{\text{concat}}^{(t)})$. 
Each step is represented as a bimanual tuple
\[
\tau_j = \bigl(
h_j^{R},\, p_j^{R},\, r_j^{R},\;
h_j^{L},\, p_j^{L},\, r_j^{L}
\bigr),
\]
where $h_j^{L}$ and $h_j^{R}$ denote the left and right hand action primitives 
(e.g., grasp, put, push, pull), which are later mapped to parameterized Unity functions.  
The interaction points $p_j^{L}$ and $p_j^{R}$ specify the selected targets, and 
$r_j^{L}$ and $r_j^{R}$ justify why each action is feasible under
the current hand state $H_t$ (which hand is free / holding an object).

A bimanual action sequence at timestep~$t$ is an arbitrary-length list of such 
tuples,
\[
T_t = (\tau_1,\, \tau_2,\, \dots,\, \tau_M),
\]
representing the complete two-handed plan produced for that step.
To generate feasible actions, IPBP consults our proposed Retrieval-Augmented 
Skill Generation module (Skill RAG). As in the refinement stage, Skill RAG retrieves two canonical, object-neutral manipulation prototypes by embedding the selected skill name $b_t$ with \texttt{all-mpnet-base-v2} and selecting the most semantically compatible entries from $L_{\text{skill}}$ in the knowledge base $K$. These retrieved prototypes serve as few-shot examples for in-context prompting, guiding the planner toward visually grounded and affordance-consistent bimanual tuples.

The retrieved prototype from $K$ is represented as
\[
k_j =
\bigl(
L_{\text{skill}},\,
C_{\text{skill}},\,
\pi_{\text{skill}},\,
P_{\text{skill}},\,
H_{\text{skill}}
\bigr),
\]
where each component provides structural and behavioral constraints for 
instantiating the skill:

\begin{itemize}

    \item \textbf{$L_{\text{skill}}$ (Canonical Skill Name):}
    A neutralized skill name describing the intended action pattern, independent 
    of any specific object.     
    \\
    \textit{Example:} ``grab two items with two hands,'' ``open a lid,'' 
    ``pour from container A to container B.''

    \item \textbf{$C_{\text{skill}}$ (Coordination Type):}
    Specifies how both hands coordinate. 
    \\
    \textit{Example:} two-handed manipulation on one object, two-handed 
    manipulation on separate objects, or strictly unimanual operation.

    \item \textbf{$\pi_{\text{skill}}$ (Canonical Bimanual Sequence Template):}
    A representative bimanual tuple sequence that acts as structural guidance 
    for IPBP's output.

    \item \textbf{$P_{\text{skill}}$ (Canonical Interaction-Point Template):}
    Example interaction-point descriptors written in the same semantic format as 
    VPG’s descriptors. These act as preconditions and provide a template for 
    matching and binding sampled interaction points.
    \\
    \textit{Example:} 
    ``Object = Small Item,\ Visible = On Table,\ Affordance = Grab,’’  
    ``Object = Container,\ Visible = Filled,\ Affordance = Pour.’’

    \item \textbf{$H_{\text{skill}}$ (Hand-Occupancy Preconditions):}
    Indicates which hands must be free or already holding an item before the 
    skill can be executed.

\end{itemize}

IPBP inserts explicit instructions into the prompt that guide the bimanual planner in applying these templates, drawing on the structured components of each prototype:

\paragraph{Enforcing proper bimanual coordination}
When the coordination type $C_{\text{skill}}$ indicates a two-handed operation, 
the prompt directs the planner to instantiate both hands in the tuple rather 
than defaulting to unnecessary unimanual steps.  
It also specifies whether the hands should operate on different parts of the 
same object or on two distinct objects, enabling appropriate part-wise 
reasoning.

\paragraph{Following the canonical action structure}
IPBP provides additional guidance instructing the planner to follow the 
canonical tuple structure and action pattern encoded in $\pi_{\text{skill}}$, 
ensuring consistency with the intended manipulation semantics.

\paragraph{Using visually compatible interaction points}
The prompt incorporates the interaction-point summary $d_{\text{concat}}^{(t)}$, 
instructing the planner to select interaction points that match the region 
descriptors $P_{\text{skill}}$ and are visually and affordance compatible.

\paragraph{Respecting hand-occupancy constraints}
When $H_{\text{skill}}$ indicates that a hand is unavailable or 
that the default left/right roles should be reversed, IPBP adds explicit 
instructions to reassign hand roles within the tuple while preserving the 
skill’s intended coordination behavior.

Along with the Skill-RAG prompt, rule-based constraints are also attached to the 
prompt to ensure physical plausibility:

\begin{itemize}
    \item \textbf{Hand-state consistency:} no re-grasping of objects already 
    held; only free hands may initiate a new grasp; objects remain with the same 
    hand until explicitly released.
    \item \textbf{Zone feasibility:} the left hand may only manipulate 
    interaction points in the \textit{left} or \textit{mid} zones, and the 
    right hand only in the \textit{right} or \textit{mid} zones.
    \item \textbf{Visual Attribute Affordance Alignment:} use visual attributes 
    in the descriptors to infer feasible hand actions, such as whether a door 
    opens horizontally or vertically and whether a component is movable or fixed.
\end{itemize}

\subsubsection{Scene-State Update and Subgoal Advancement}
After generating $T_t$, the scene-update module updates the object state, hand state, and completed subgoals within a single prompt.
This results in the transition
\[
\bigl(
H_{t+1},\,
d_{\text{concat}}^{(t+1)},\,
G_{\text{c}}^{t+1}
\bigr)
=
f_{\text{update}}\!\left(
H_t,\,
d_{\text{concat}}^{(t)},\,
T_t,\,
g_{t},\,
G_{\text{c}}^{t}
\right).
\]

Here, the updated hand state $H_{t+1}$ is inferred using the chain of
reasons $r_j^{L},~ r_j^{R}$ together with the unfolding hand operators
$h_j^{L},~ h_j^{R}$ in $T_t$. The descriptor $d_{\text{concat}}^{(t)}$ is updated
to reflect the new object state in text form. $G_{\text{target}}$ is satisfied, it is added to $G_{\text{c}}$, which stores the list of
completed goals.

\subsection{Parameterized Manipulation Operators}

Each tuple $\tau_j$ in $T_t$ is executed through parameterized low-level 
manipulation operators. Each operator takes an interaction point 
$p = (x, d)$ and performs a predefined action such as 
grasping, placing, pressing, pulling, or releasing.

For each tuple $\tau_j$, the planner invokes both the right- and left-hand 
operators corresponding to $h_j^{R}$ and $h_j^{L}$, supplying the interaction 
point parameters $p_j^{R}$ and $p_j^{L}$. Both operators are executed 
synchronously to produce a coordinated bimanual motion. After execution, the 
scene and hand-state descriptors are updated before proceeding to the next tuple 
in $T_t$.

\section{Evaluation}

We evaluate our framework with three objectives:
(1) determining whether it generates valid, constraint-grounded bimanual behaviors;
(2) measuring whether it produces compact plans requiring fewer action tuples; and
(3) examining its ability to generalize across different scene and activity domains in a zero-shot manner.

All experiments are conducted in a Unity-based simulator using RootMotion’s Final IK for hand-motion execution and an object-interaction system for contact-based operations such as pressing, grasping, and releasing. Feasible placement locations are computed via object-point--based raycasting. All language and vision--language modules use GPT-4.1.

\subsection{Evaluation Setup}

We report two quantitative metrics as follows:

\textbf{Success Rate (Succ.):} A trial is considered successful if the agent completes the task without violating feasibility or affordance constraints. The system flags errors such as manipulating the wrong object, using an incorrect hand, or attempting an action inconsistent with object affordances. The success rate reflects the fraction of runs completed without violations.

\textbf{Operation Count (Op.): } Operation Count is the total number of bimanual action tuples used to complete a successful trial. Each tuple corresponds to one synchronous left--right action step. Lower Op.\ values indicate more compact plans.

\begin{table}[t]
  \caption{Success rate and operation count per scene--task pair (20 trials each).}
  \vspace{-5pt}
  \footnotesize
  \label{tab:subactivity_results}
  \centering
  \begin{tabular}{lcc}
    \toprule
    Scene--Task  & Succ.  & Op. \\
    \midrule
    Outdoor Yard / Throw Away a Trash     & 20/20  & 2.00 \\
    Outdoor Yard / Water Flower           & 19/20  & 2.16 \\
    Convenience Store / Buy Two Bottles of Coke & 19/20 & 2.00 \\
    Convenience Store / Heat a Lunch Box  & 18/20  & 3.06 \\
    Cafe / Pour a Tea                     & 17/20  & 3.00 \\
    Cafe / Make a Coffee                  & 20/20  & 2.00 \\
    \midrule
    Average                                & 93.33\%& 2.36 \\
    \bottomrule
  \end{tabular}
\end{table}

\subsection{Generalization Across Scene--Task Domains}

We evaluate the system across three distinct 3D environments without any 
retraining or parameter adjustment. Generalization is demonstrated by consistent,
constraint-feasible plans across:

\begin{itemize}
    \item Outdoor Yard: Throw Away a Trash; Water a Flower
    \item Convenience Store: Buy Two Bottles of Coke; Heat a Lunch Box
    \item Cafe/Kitchen: Pour a Tea; Make a Coffee
\end{itemize}

\subsection{Component Ablations}

To validate the roles of the bimanual merge rule and annotated-skill retrieval scheme, 
we conduct ablations by disabling one component at a time and measuring its effect on 
success rate and operation count.

\subsubsection{w/o VPG (No visual affordance, attribute, or part-wise reasoning)}
In this setting, all reasoning—affordances, visual attributes, within-object relations, and part-wise cues—is removed. To keep the pipeline minimally workable, the VLM returns only the parent object name and object state, and each part is assigned a default empty hand state with no further interpretation.

\subsubsection{w/o BSP (No Merging)}
This ablation removes the adjacency graph and the bimanual merge rule.
Subgoals are generated independently at each object point—without checking joint
reachability, merging adjacent points, or resolving continuity placeholders (e.g., “while holding~A”).

\subsubsection{w/o Skill RAG}
IPBP does not retrieve any canonical skill templates from the knowledge base as few-shot examples when generating bimanual sequences. Instead, the prompt includes only hand-state consistency, zone feasibility, and visual-attribute affordance alignment constraints.

\begin{table}[t]
  \caption{Ablation study of framework components over 120 trials 
  (6 subactivities × 20 trials).}
  \vspace{-5pt}
  \label{tab:ablations}
  \centering
    \begin{tabular}{lcc}
    \toprule
    Variant & Succ.~(\%, $\uparrow$) & Op. $\downarrow$ \\
    \midrule
    Full (ours)         & \textbf{93.33} & \textbf{2.36} \\
    w/o BSP             & 71.66 & 2.82 \\
    w/o Skill RAG & 18.33 & 3.08 \\
    w/o VPG & 0.00 & 0.00 \\
    \bottomrule
    \end{tabular}
    \vspace{-5pt}
\end{table}

\begin{figure}[t]
    \centering
    \includegraphics[width=0.885\columnwidth]{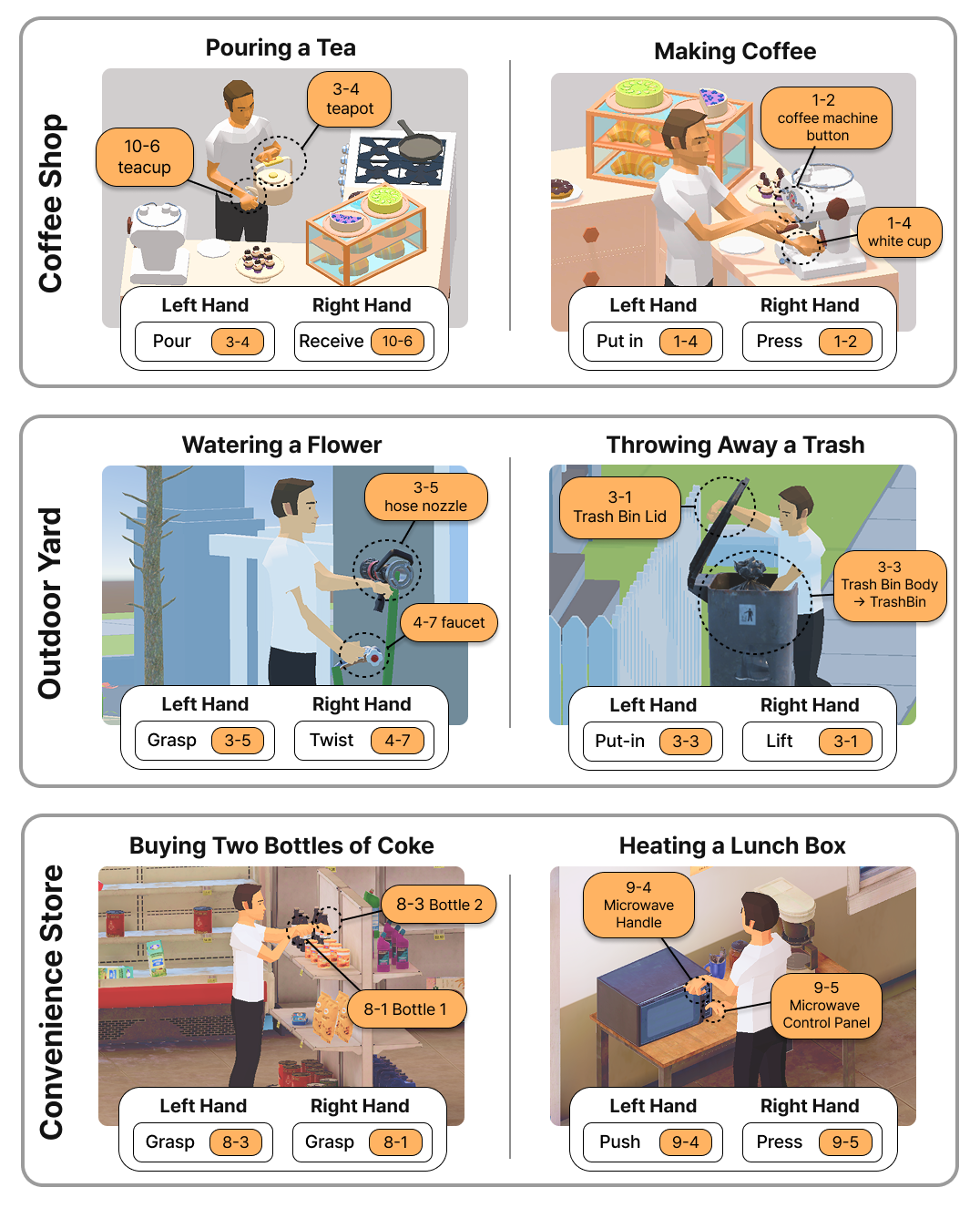}
    \vspace{-10pt}
    \caption{Overview of bimanual action tuples generated by the 
    Interaction-Point--Driven Bimanual Prompting system across diverse scene--task pairs. 
    Each example shows a synchronized left--right action grounded in local 
    interaction-point identifiers, demonstrating consistent, affordance-aligned 
    bimanual behavior despite variations in scene layout and object configuration.}
    \label{fig:figure3}
    \vspace{-10pt}
\end{figure}

\subsection{Results and Discussion}

Table~\ref{tab:subactivity_results} summarizes the repeated-trial performance.  
The framework achieves an average success rate of 93.33\%, demonstrating reliable generation of bimanual behaviors grounded in scene geometry and affordances. Performance remains consistently high across all scene–task combinations, indicating strong robustness to unseen environments. The system retrieves task-relevant objects and extracts interaction points in a zero-shot manner, producing visually aligned and affordance-consistent bimanual manipulations that support scene-agnostic execution. As shown in Figure~\ref{fig:figure3}, the resulting action tuples remain spatially coherent with each task across varied environments.

In failure cases, LLMs exhibit characteristic hallucination patterns in bimanual reasoning.  
Multi-step subgoals can become entangled, causing the model to insert or modify intermediate steps or drift toward unimanual execution.  
Tasks requiring asymmetric hand roles may trigger inversions of source--target semantics.

The ablation study, summarized in Table~\ref{tab:ablations}, clarifies the role of each module.
Removing BSP reduces success to 71.66\% and increases Op.\ to 2.82.
Without the merge rule, subgoals remain fragmented and redundant, preventing the planner from exploiting natural bimanual opportunities.
Continuity placeholders such as ``while holding~A'' also fail to resolve because no merge-aware logic binds them to their originating grasp step, resulting in inefficient and spatially inconsistent plans.

Removing Skill-RAG yields severe degradation (18.33\% Succ., Op.\ 3.08).
Without canonical templates supplying preconditions, affordance cues, or coordination patterns, the LLM must infer all bimanual roles from generic hand-state constraints alone.
This leads to frequent errors, including both hands selecting the same object, loss of hand–object continuity, redundant re-grasps, and failure on asymmetric tasks such as pouring or insertion while supporting.

Ablating VPG drops the success rate to 0\%, underscoring its essential role in translating scene-level visual cues into grounded bimanual planning. Without visual attributes, the agent violates basic physical constraints—for example, attempting to grasp wall-mounted fixtures such as faucets that cannot be relocated, or failing to determine how a door should be opened, whether it is a horizontally hinged door to pull or a vertically opening door to lift. Removing part-wise affordance reasoning further prevents the agent from distinguishing object components, leading to errors such as lifting a trash bin’s body instead of its lid. In some cases, it also grasps an object with two hands unnecessarily because it fails to recognize that multiple interaction points belong to the same object.

Overall, the results validate the three evaluation objectives.  
High success rates show that the generated behaviors satisfy feasibility and 
affordance constraints.  
Low operation counts demonstrate that explicit modeling of bimanual coordination 
and adjacency yields compact plans.  
Consistent performance across three distinct scenes without retraining shows robust 
zero-shot generalization when object points and annotated skills are available.  
Remaining errors occur primarily in longer, multi-stage routines, suggesting future 
work on global consistency checks or lightweight plan-repair mechanisms to stabilize 
intermediate states across subgoal boundaries.
\section{Conclusion}

Our experiments demonstrate that the proposed framework reliably generates valid 
bimanual behaviors, produces compact action sequences, and generalizes across 
diverse scenes without retraining. The ablations show that each module—VPG for 
spatial grounding, BSP for object-point–level subgoal structure, and IPBP for 
executable two-handed actions—is essential to achieving these results. Remaining 
errors arise primarily in multi-step routines, suggesting future work on 
stronger consistency checks across sequential planning and execution stages.

\balance
\bibliographystyle{IEEEtran}
\bibliography{reference}

\end{document}